\def\year{2018}\relax
\documentclass[letterpaper]{article} 
\usepackage{aaai18}  
\usepackage{times}  
\usepackage{helvet}  
\usepackage{courier}  
\usepackage{url}  
\usepackage{graphicx}  
\usepackage{booktabs}
\usepackage{amsmath}
\usepackage{multicol}
\usepackage[inline]{enumitem}
\begin{document}
%
\title{Multi-Label Classification of Patient Notes: Case Study on ICD Code Assignment}

\author{Tal Baumel\\
Ben-Gurion University\\
Beer-Sheva, Israel\\
\And
Jumana Nassour-Kassis\\
Ben-Gurion University\\
Beer-Sheva, Israel\\
\And
Raphael Cohen\\
Chorus.ai\\
San Francisco, CA\\
\And
Michael Elhadad\\
Ben-Gurion University\\
Beer-Sheva, Israel\\
\And
No\'emie Elhadad\\
Columbia University\\
New York, NY\\
}

\maketitle
\begin{abstract}
The automatic coding of clinical documentation according to diagnosis codes is a useful task in the Electronic Health Record, but a challenging one due to the large number of codes and the length of patient notes. We investigate four models for assigning multiple ICD codes to discharge summaries, and experiment with data from the MIMIC II and III clinical datasets. We present Hierarchical Attention-bidirectional Gated Recurrent Unit (HA-GRU), a hierarchical approach to tag a document by identifying the sentences relevant for each label. HA-GRU achieves state-of-the art results. Furthermore, the learned  sentence-level attention layer highlights the model decision process, allows for easier error analysis, and suggests future directions for improvement.
\end{abstract}
\section{Introduction}

In Electronic Health Records (EHRs), there is often a need to
assign multiple labels to a patient record, choosing from a large number
of potential labels. Diagnosis code assignment is such a task,
with a massive amount of labels to chose from (14,000 ICD9 codes and
68,000 ICD10 codes). Large-scale multiple phenotyping assignment,
problem list identification, or even intermediate patient representation
can all be cast as a multi-label classification over a large label
set. More recently, in the context of predictive modeling, approaches to
predict multiple future healthcare outcomes, such as future diagnosis
codes or medication orders have been proposed in the literature. There
again, the same setup occurs where patient-record data is fed to a multi-label classification over a large label set.

In this paper, we investigate how to leverage the unstructured portion
of the EHR, the patient notes, along a novel application of neural architectures. We focus on three characteristics: \textbf{(i) a very large label set} (6,500 unique ICD9 codes and 1,047 3-digit unique ICD9 codes); \textbf{(ii) a multi-label setting} (up to 20 labels per instance); \textbf{(iii) instances are long documents} (discharge summaries on average 1,900-word long); and \textbf{(iv)} furthermore, because we work on long documents, one critical aspect of the multi-label classification is \textbf{transparency}---to highlight the elements in the documents that explain and support the predicted labels.  While there has been much work on each of these characteristics, there has been limited work to tackle all at once, particularly in the clinical domain. 

We experiment with four approaches to classification: an \textbf{SVM-based one-vs-all} model, \textbf{a continuous bag-of-words} (CBOW) model, \textbf{a convolutional neural network} (CNN) model, and \textbf{a bidirectional Gated Recurrent Unit model with a Hierarchical Attention mechanism} (HA-GRU). Among them, the attention mechanism of the HA-GRU model provides full \textbf{transparency for classification decisions}. We rely on the publicly available MIMIC datasets to validate our experiments. A characteristic of the healthcare domain is long documents with a large number of technical words and typos/misspellings. We experiment with simple yet effective preprocessing of the input texts. 

Our results show that careful tokenization of the input texts, and hierarchical segmentation of the original document allow our Hierarchical Attention GRU architecture to yield the most promising results, over the SVM, CBOW, and CNN models, while preserving the full input text and providing effective transparency. 

\section{Previous Work}
We review previous work in the healthcare domain as well as recent approaches to extreme multi-label classification, which take place in a range of domains and tasks.

\subsection{Multi-label Patient Classifications}
Approaches to classification of patient records against multiple labels fall into three types of tasks: diagnosis code assignment, patient record labeling, and predictive modeling.

\paragraph*{Diagnosis Code Assignment.}
Automated ICD coding is a well established task, with several methods proposed in the literature, ranging from rule based \cite{crammer2007automatic,farkas2008automatic} to machine learning such as support vector machines, Bayesian ridge regression, and K-nearest neighbor \cite{larkey1995automatic,lita2008large}. Some methods exploit the hierarchical structure of the ICD taxonomy~\cite{perotte2011hierarchically,perotte2014diagnosis}, while others incorporated explicit co-occurrence relations between codes~\cite{kavuluru2015empirical}. In many cases, to handle the sheer amount of labels, the different approaches focus on rolled-up ICD codes (i.e., 3-digit version of the codes and their descendants in the ICD taxonomy) or on a subset of the codes, like in the shared community task for radiology code assignment ~\cite{pestian2007shared}.

It is difficult to compare the different methods proposed, since each relies on different (and usually not publicly available) datasets. We experiment with the MIMIC dataset, since it is publicly available to the research community. Methods-wise, our approach departs from previous work in two important ways: we experiment with both massively large and very large label sets (all ICD9 code and rolled-up ICD9 codes), and we experiment with transparent models that highlight portions of the input text that support the assigned codes.

\paragraph*{Patient Record Labeling.} Other than automated diagnosis coding, most multi-label patient record classifiers fall in the tasks of phenotyping across multiple conditions at once. For instance, the UPhenome model takes a probabilistic generative approach to assign 750 latent variables~\cite{pivovarov2015learning}. More
recently, in the context of multi-task learning, Harutyunyan and colleagues experimented with phenotyping over 25 critical care 
conditions~\cite{harutyunyan2017multitask}. 

\paragraph*{Predictive Modeling.} Previous work in EHR multi-label classification has mostly focused on predictive scenarios. The size of the label set
varies from one approach to another, and most limit the label set size however: 
DeepPatient~\cite{miotto2016deep} predicts over a set of 78 condition codes. \cite{lipton2015learning} leverage an LSTM model to predict over a vocabulary of 128 diagnosis codes. DoctorAI~\cite{choi2015doctor} predicts over a set of 1,183 3-digit ICD codes and 595 medication groups. The
Survival Filter~\cite{ranganath2015survival} predicts a series of future ICD codes across approximately 8,000 ICD codes.

\paragraph*{Inputs to Multi-Label Classifications.} 
Most work in multi-label classification takes structured input. For instance, the Survival Filter expects ICD codes as input to predict the future ICD codes. DoctorAI takes as input medication orders, ICD codes, problem list, and procedure orders at a given visit. Deep Patient does take the content of notes as input, but the content is heavily preprocessed into a structured input to their neural network, by tagging all texts with medical named entities. In contrast, our approach is to leverage the entire content of the input texts. Our work contributes to clinical natural language processing~\cite{demner2016aspiring}, which only recently investigated neural representations and architectures for traditional tasks such as named entity recognition~\cite{jagannatha2016structured}.

\subsection{Multi-label Extreme Classification}
In extreme multi-label learning, the objective is to annotate each data point with the most relevant subset of labels from an extremely large label set. 
Much work has been carried outside of the healthcare domain on tasks such as image  classification~\cite{tsoumakas2006multi,weston2011wsabie}, question answering~\cite{choi2016hierarchical}, and advertising~\cite{jain2016extreme}.
In \cite{weston2011wsabie}, the task of annotating a very large dataset of images ($>10M$) with a very large label set ($>100K$) was first addressed. The authors introduced the WSABIE method which relies on two main features: (i) records (images) and labels are embedded in a shared low-dimension vector space; and (ii) the multi-label classification task is modeled as a ranking problem, evaluated with a Hamming Loss on a P@k metric.  The proposed online approximate WARP loss allowed the algorithm to perform fast enough on the scale of the dataset.  We found that in our case, the standard Micro-F measure is more appropriate as we do not tolerate approximate annotations to the same extent as in the image annotation task.

The SLEEC method \cite{bhatia2015sparse} also relies on learning an embedding transformation to map label vectors into a low-dimensional representation.  SLEEC learns an ensemble of local distance preserving embeddings to accurately predict infrequently occurring labels. This approach attempts to exploit the similarity among labels to improve classification, and learns different representations for clusters of similar labels. Other approaches attempt to reduce the cost of training over very large datasets by considering only part of the labels for each classification decision \cite{yen2016pd}. SLEEC was later improved in \cite{jain2016extreme} with the PfastreXML method which also adopted P@k loss functions aiming at predicting tail labels. 

In \cite{joulin2016fasttext}, the FastText method was introduced as a simple and scalable neural bag of words approach for assigning multiple labels to text.  We test a similar model (CBOW) in our experiments as one of our baselines.

\section{Dataset and Preprocessing}

We use the publicly available de-identified MIMIC dataset of ICU stays from Beth Israel Deaconess Medical Center~\cite{saeed2011multiparameter,johnson2016mimic}. 

\subsection{MIMIC Datasets}

To test the impact of training size, we relied on both the MIMIC II (v2.6) and MIMIC III (v1.4) datasets. MIMIC III comprises records collected between 2001 and 2012, and can be described as an expansion of MIMIC II (which comprises records collected between 2001 and 2008), along with some edits to the dataset (including de-identification procedures). 

To compare our experiments to previous work in ICD coding, we used the publicly available split of MIMIC II from ~\cite{perotte2014diagnosis}. It contains 22,815 discharge summaries divided into a training set (20,533 summaries) and a test-set of unseen patients (2,282 summaries).  We thus kept the same train and the test-set from MIMIC II, and constructed an additional training set from MIMIC III. We  made sure that the test-set patients remained unseen in this training set as well. Overall, we have two training sets, which we refer to as MIMIC II and MIMIC III, and a common test-set comprising summaries of unseen patients. 


While there is a large overlap between MIMIC II and MIMIC III, there are also marked differences. We found many cases where discharge summaries from 2001-2008 are found in one dataset but not in the other. In addition, MIMIC III contains addenda to the discharge summaries that were not part of MIMIC II.  After examining the summaries and their addenda, we noticed that the addenda contain vital information for ICD coding that is missing from the main discharge summaries; therefore, we decided to concatenate the summaries with their addenda.

Table~\ref{tab:stats} reports some descriptives statistics regarding the datasets. Overall, MIMIC III is larger than MIMIC II from all standpoints, including amounts of training data, vocabulary size, and overall number of labels.   



\begin{table}
 \centering
 \begin{small}
    \begin{tabular}{lrrr}
    \toprule
    & MIMIC II & MIMIC III & Test Set \\
    \midrule
    Nb of records & 20,533 & 49,857 & 2,282\\
    Nb of unique tokens & 69,248 & 119,171 & 33,958\\
    Avg nb of tokens / record & 1,529 & 1,947 & 1,893\\
    Avg nb of sentences / record & 90  & 112 &  104 \\
    Nb of full labels & 4,847& 6,527 & 2,451\\
    Nb of rolled-up labels & 948 & 1,047 & 684 \\
    Label Cardinality & 9.24 & 11.48 & 11.42\\
    Label Density & 0.0019 & 0.0018 & 0.0047\\
    \% labels with 50+ records & 11.33\%& 18.19\%& 4.08\%\\
    \bottomrule 
    \end{tabular}
    \end{small} 
     \caption{Datasets descriptive statistics.}
     \label{tab:stats}
    \end{table}
    
\subsection{ICD9 Codes}
Our label set comes from the ICD9 taxonomy. The 
International Classification of Diseases (ICD) is a repository maintained by the World Health Organization (WHO) to provide a standardized system of diagnostic codes for classifying diseases. It has a hierarchical structure, connecting specific diagnostic codes through is-a relations. The hierarchy has eight levels, from less specific to more specific. ICD codes contain both diagnosis and procedure codes. In this paper, we focus on diagnosis codes only. ICD9 codes are conveyed as 5 digits, with 3 primary digits and 2 secondary ones.

Table~\ref{tab:stats} provides the ICD9 label cardinality and density as defined by~\cite{tsoumakas2006multi}. Cardinality is the average number of codes assigned to records in the dataset. Density is the cardinality divided by the total number of codes. For both training sets, the number of labels is of the same order as the number of records, and the label density is extremely low. This confirms that the task of code assignment belongs to the family of extreme multi-label classification.

We did not filter any ICD code based on their frequency. We note, however that there are approximately 1,000 frequent labels (defined as assigned to at least 50 records) (Table~\ref{tab:stats}). We experimented with two versions of the label set: one with all the labels (i.e., 5-digit), and one with the labels rolled up to their 3-digit equivalent. 

\subsection{Input Texts}

\paragraph*{Tokenization.}
Preprocessing of the input records comprised  the following steps: (i) tokenize all input texts using spaCy library \footnote{https://spacy.io/}; (ii) convert all non-alphabetical characters to pseudo-tokens (e.g., ``11/2/1986'' was mapped to ``dd/d/dddd''); (iii) build the vocabulary as tokens that  appear at least 5 times in the training set; and (iv) map any out-of-vocabulary word to its nearest word in the vocabulary (using the edit distance). This step is simple, yet particularly useful in reducing the number of misspellings of medical terms.  
These preprocessing steps has a strong impact on the vocabulary. For instance, there were 1,005,489 unique tokens in MIMIC III and test set before preprocessing, and only 121,595 remaining in the vocabulary after preprocessing (an 88\% drop). This step improved F-measure performance by \textasciitilde0.5\% when tested on the CBOW and CNN methods (not reported).

\paragraph*{Hierarchical Segmentation.}
Besides tokenization of the input texts, we carried one more level of segmentation, at the sentence level (using the spaCy library as well). There are two reasons for preprocessing the input texts with sentence segmentation. 
First, because we deal with long documents, it is impossible and ineffective to train a sequence model like an GRU on such long sequences. In previous approaches in document classification, this problem was resolved by truncating the input documents. In the case of discharge summaries, however, this is not an acceptable solution: we want to preserve the entire document for transparency. Second, we are inspired by the moving windows of ~\cite{johnson2014effective} and posit that sentences form linguistically inspired windows of word sequences. 

Beyond tokens and sentences, discharge summaries exhibit strong discourse-level structure (e.g., history of present illness and past medical history, followed by hospital course, and discharge plans)~\cite{li2010section}. This presents an exciting opportunity for future work to exploit discourse segments as an additional representation layer of input texts.

\section{Methods}
We describe the four models we experimented with. 
ICD coding has been evaluated in the literature according to  different metrics: Micro-F, Macro-F, a variant of Macro-F that takes into account the hierarchy of the codes \cite{perotte2014diagnosis}, Hamming and ranking loss \cite{wang2016diagnosis}, and a modified version of mean reciprocal rank (MRR) \cite{subotin2014system}. We evaluate performance using the Micro-F metric, since it is the most commonly used metric.

\paragraph*{SVM.} 
We used Scikit Learn~\cite{pedregosa2011scikit} to implement a one-vs-all, multi-label binary SVM classifier. Features were bag of words, with tf*idf weights (determined from the corpus of release notes) for each label. Stop words were removed using Scikit Learn default English stop-word list. The model fits a binary SVM classifier for each label (ICD code) against the rest of the labels.  We also experimented with $\chi^2$ feature filtering to select the top-N words according to their mutual information with each label, but this did not improve performance.

\paragraph*{CBOW.}
The continuous-bag-of-words (CBOW) model is inspired by the word2vec CBOW model \cite{mikolov2013efficient} and FastText \cite{joulin2016fasttext}. Both methods use a simple neural-network to create a dense representation of words and use the average of this representation for prediction. The word2vec CBOW tries to predict a word from the words that appear around it, while our CBOW model for ICD classification predicts ICD9 codes from the words of its input discharge summary.

\begin{figure}
   \centering
   \includegraphics[width=2.2in]{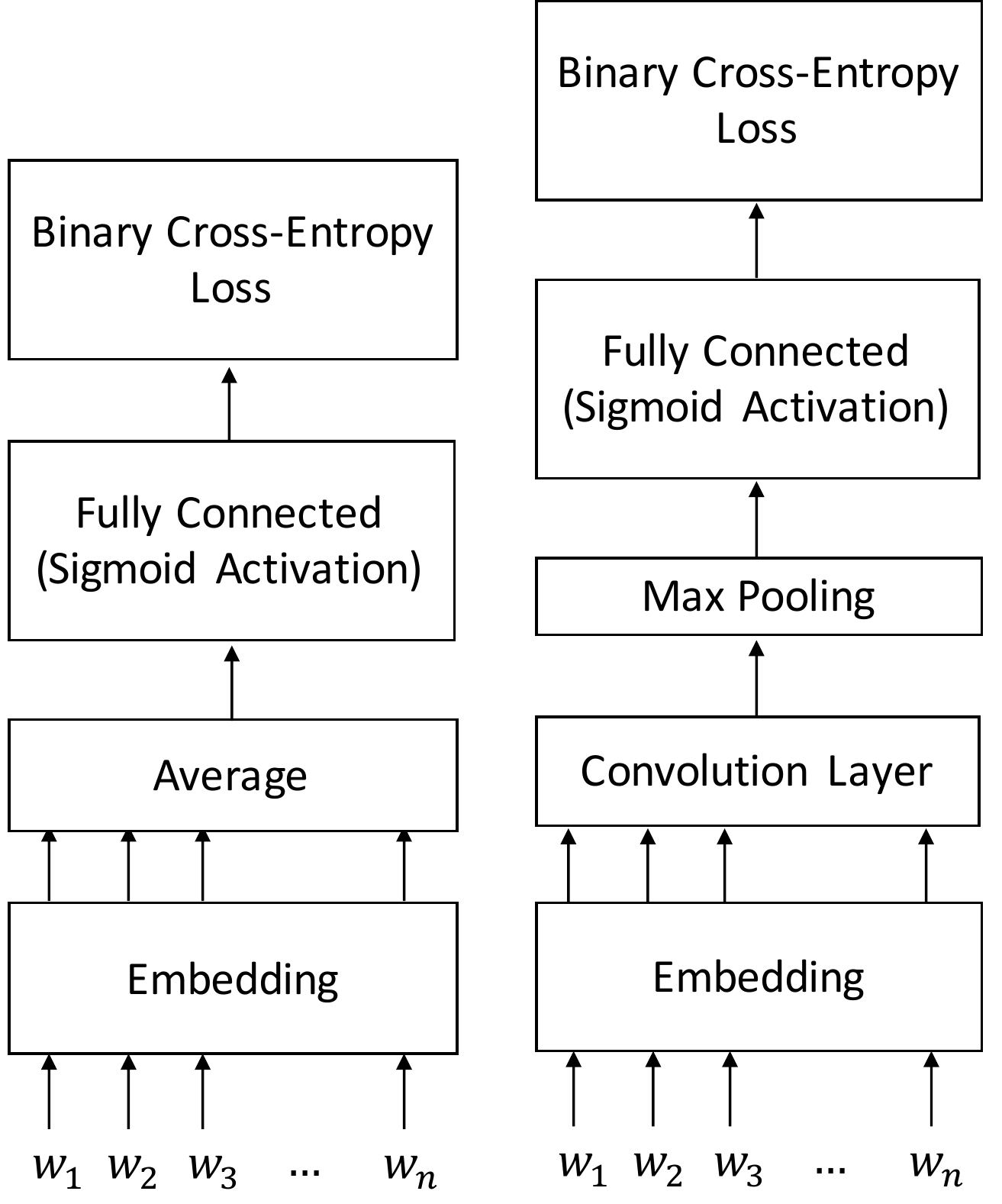}
   \caption{CBOW architecture on the left and CNN model architecture on the right.}\label{fig:myfig}
\end{figure}

The model architecture consists of an embedding layer applied to  all the words in a given input text $[w_1,w_2,...,w_n]$, where $w_i$ is a one-hot encoding vector of the vocabulary. $E$ is the embedding matrix with dimension $n_{emb} \times V$, where $V$ is the size of the vocabulary and $n_{emb}$ is the embedding size (set to 100).

The embedded words are averaged into a fixed-size vector and are fed to a fully connected layer with a matrix $W$ and bias $b$, where the output dimension is the number of labels. We use a sigmoid activation on the output layer so all values are in the range of $[0-1]$ and use a fixed threshold (0.5) to determine whether to assign a particular label. To train the model, we used binary cross-entropy loss 
($loss(target, output)=-(target\cdot \log(output)+(1-target)\cdot\log(1-output)$).

\begin{tiny}
\begin{align*}
Embedding &= E\cdot[w_1,w_2,...,w_n]\\
Averaged &= 1/n\Sigma_{e\in Embedding}(e)\\
Prob &= sigmoid(W \cdot Averaged + b)
\end{align*}
\end{tiny}
While the model is extremely lightweight and fast it suffers from known bag-of-words issues: (i) it ignores word order; i.e., if negation will appear before a diagnosis mention, the model would not be able to learn this; (ii) multi-word-expressions cannot be identified by the model, so different diagnoses that share lexical words will not be distinguished by the model.

\paragraph*{CNN.}
To address the problems of the CBOW model, the next model we investigate is a convolutional neural network (CNN). A one dimensional convolution applied on list of embedded words could be considered as a type of n-gram model, where n is the convolution filter size.

The architecture of this model is very similar to the CBOW model, but instead of averaging the embedded words we apply a one dimensional convolution layer with filter $f$, followed by a max pooling layer. On the output of the max pool layered a fully connected layer was applied, like in the CBOW model. We also experimented with deeper convolution networks and inception module \cite{lecun2015lenet}, but they did not yield improved results. 

\begin{tiny}
\begin{align*}
Embedding &= E\cdot[w_1,w_2,...,w_n]\\
Conved &= \max_{i\in channels}(Embedding\ast f)\\
Prob &= sigmoid(W \cdot Conved + b)
\end{align*}
\end{tiny}
In our experiments, we used the same embedding parameter as in the CBOW model. In addition, we set the number of channels to 300, and the filter size to 3. 

\paragraph*{HA-GRU.}
We now introduce the Hierarchical Attention-bidirectional Gated Recurrent Unit model (HA-GRU) an adaptation of a Hierarchical Attention Networks~\cite{yang2016hierarchical} to be able to handle multi-label classification. A Gated Recurrent Unit (GRU) is a type of Recurrent Neural Network. Since the documents are long (see Table~\ref{tab:stats} ad up to 13,590 tokens in the MIMIC III training set), a regular GRU applied over the entire document is too slow as it requires number layers of the document length. Instead we apply a hierarchal model with two levels of bidirectional GRU encoding. The first bidirectional GRU operates over tokens and encodes sentences. The second bidirectional GRU encodes the document , applied over all the encoded sentences. In this architecture,  each GRU is applied to a much shorter sequence compared with a single GRU. 

To take advantage of the property that each label is invoked from different parts of the text, we use an attention mechanism over the second GRU with different weights for each label. This allows the model to focus on the relevant sentences for each label~\cite{choi2016hierarchical}. To allow clarity into what the model learns and enable error analysis attention is also applied over the first GRU with the same weights for all the labels.
\begin{figure}[htbp]
  \centering 
  \includegraphics[width=3.4in]{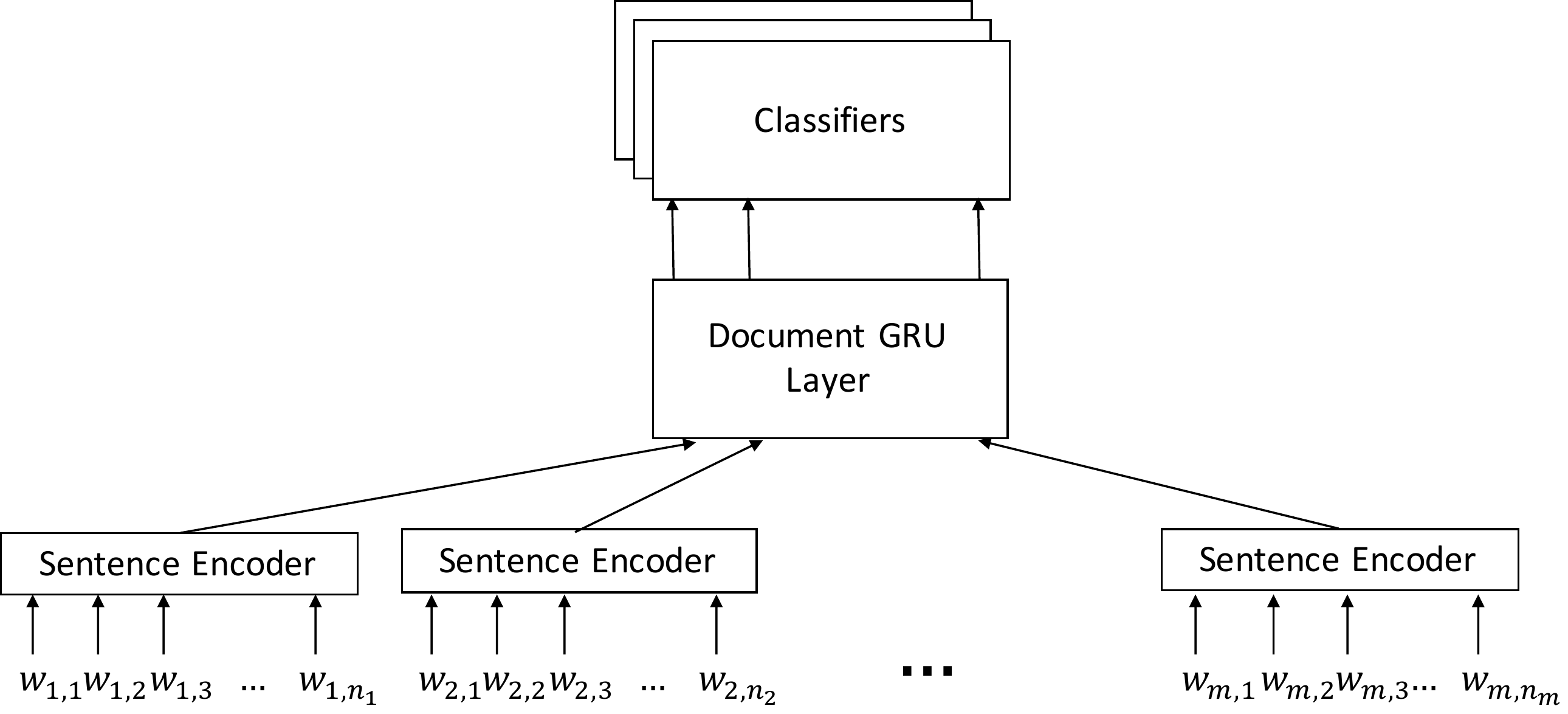} 
  \caption{HA-GRU model architecture overview.}\label{fig:HA-GRU} 
\end{figure} 

Each sentence in the input text is encoded to a fixed length vector (64) by applying an embedding layer over all the inputs, applying a bidirectional GRU layer on the embedded words, and using a neural attention mechanism to encode the bidirectional GRU outputs (size of 128). After the sentences are encoded into a fixed length vector, we apply a second bidirectional GRU layer over the sentences using different attention layers to generate an encoding specified to each class ($128\times\#labels$). Finally we applied a fully connected layer with softmax for each classifier to determine if the label should be assigned to the document. Training is achieved by using categorical cross-entropy on every classifier separately ($loss(target, output)=-\sum_x ouput(x)\cdot log(target(x))$)

\begin{tiny}
\begin{align*}
AttWeight(in_i,v,w) &= v\cdot tanh(w\cdot (in_i))\\
\overline{AttWeight}(in_i,v,w) &=\frac{e^{AttWeight(in_i,v,w)}}{e^{\sum_j AttWeight_j(v,w)}}\\ 
Attend(in,v,w) &= sum(in_i\cdot \overline{AttWeight}(in_i,v,w))\\
Embedding &= E\cdot[w_1,w_2,...,w_n]\\
EncSents_j &= Attend(GRU{words}(Embedding),v_{words},w_{words})\\
EncDoc_{label} &= Attend(GRU_{sents}(EncSents,v_{label},w_{label}),)\\
Prob_{label} &= softmax(pw_{label} \cdot EncDoc_{label} + pb_{label})
\end{align*}
\end{tiny}
Where $w_i$ is a one-hot encoding vector of the vocabulary size $V$, $E$ is an embedding matrix size of $n_{emb} \times V$, $GRU_{words}$ is a GRU layer with state size $h_{state}$, $w_{words}$ is a square matrix ($h_{state} \times h_{state}$) and $v_{words}$ is a vector ($h_{state}$) for the sentence level attention. $GRU_{sents}$ is a GRU layer with state size of $h_{state}$. $w_{label}$ is a square matrix ($h_{state} \times h_{state}$) and $v_{label}$ is a vector ($h_{state}$) for the document level attention for each class, $pw_{label}$ is a matrix ($h_{state} \times 2$) and $pb_{label}$ is a bias vector with a size of $2$ for each label. We implemented the model using DyNet \cite{neubig2017dynet}\footnote{Code available at \url{https://github.com/talbaumel/MIMIC}.}.

\begin{figure}[htbp]
  \centering 
  \includegraphics[width=2.4in]{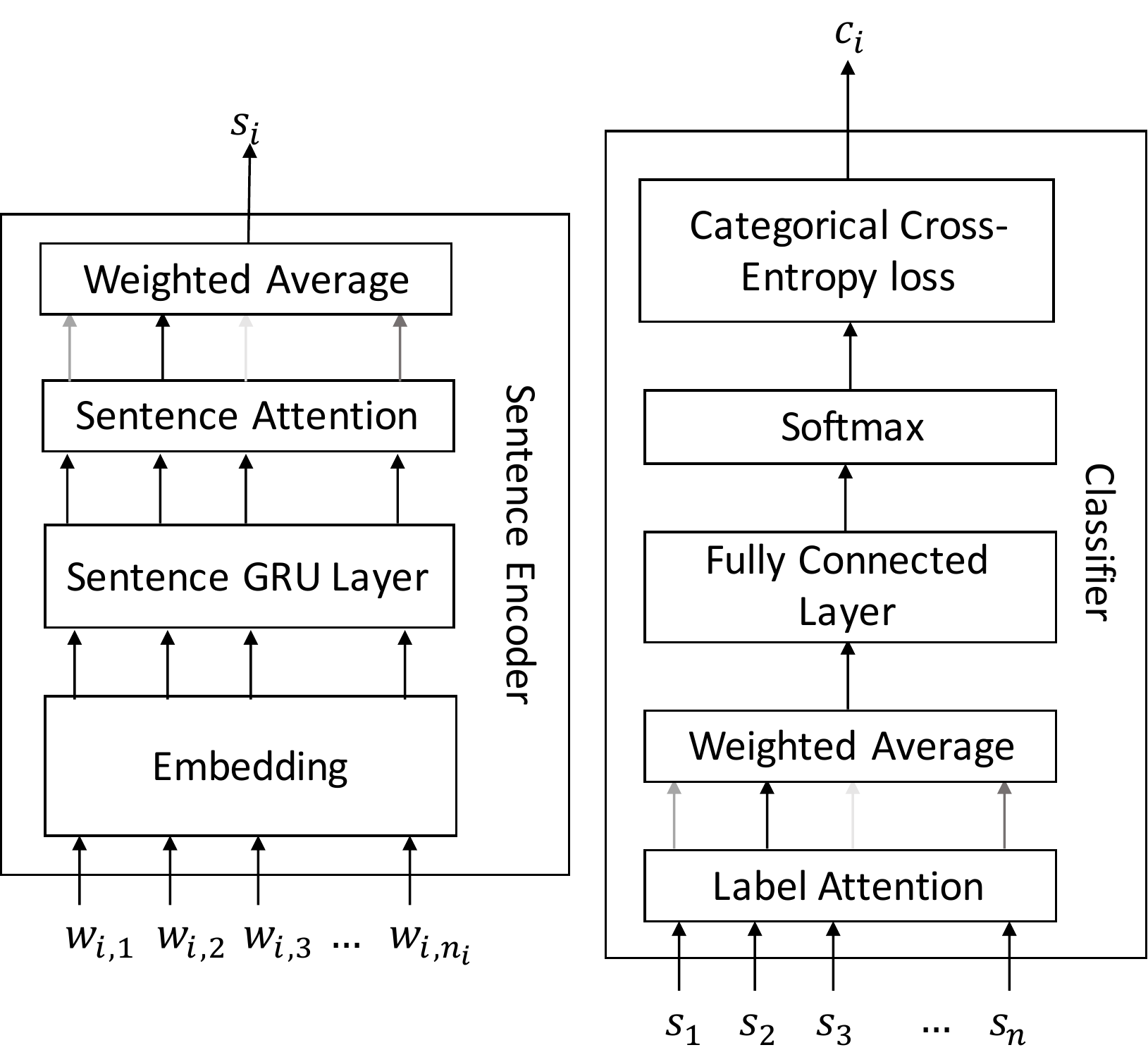} 
  \caption{Zoom-in of the sentence encoder and classifier.}
  \label{fig:zoom} 
\end{figure}

\section{Results}
\subsection{Model Comparison}
To evaluate the proposed methods on the MIMIC datasets, we conducted the following experiments. In the first setting we considered all ICD9 codes as our label set. We trained the SVM, CBOW, and CNN on the MIMIC II and on the MIMIC III training sets separately. All models were evaluated on the same test set according to Micro-F. In the second setting, we only considered the rolled-up ICD9 codes to their 3-digit codes. There (Table~\ref{results}). 

HA-GRU gives the best results in the rolled-up ICD9 setting, with a 7.4\% and  3.2\% improvement over the CNN and SVM, the second best methods, in MIMIC II and MIMIC III respectively. In the full ICD-9 scenario, all methods yield better results when trained on MIMIC III rather than on MIMIC II. This is expected considering the larger size of MIMIC III over II. We note that our CNN yields the best Micro-F when trained on MIMIC III passing the HA-GRU by a small margin.

In comparison to the previous work of  ~\cite{perotte2014diagnosis}, our one-vs-all SVM yielded better results than their flat and hierarchy classifiers. This trend was confirmed when training on the new MIMIC III set, as well as when using the same evaluation metrics of ~\cite{perotte2014diagnosis}. We attribute these improved results both to the one-vs-all approach as well as our tokenization approach. 

\begin{table}
 \centering
 \begin{small}
    \begin{tabular}{l|cc|cc}
    \toprule
     & \multicolumn{2}{c|}{ICD9 codes}  & \multicolumn{2}{c}{Rolled-up ICD9 codes} \\ 
    & MIMIC II & MIMIC III &  MIMIC II & MIMIC III \\
    \midrule
    SVM & 28.13\% & 22.25\% & 32.50\% & 53.02\%\\
    CBOW & 30.60\% & 30.02\% & 42.06\% & 43.30\%\\
    CNN & 33.25\% & \textbf{40.72\%} & 46.40\% & 52.64\%\\
    HA-GRU & \textbf{36.60\%} & 40.52\% & \textbf{53.86\%} & \textbf{55.86\%}\\
    \bottomrule 
    \end{tabular}
  \end{small}
 \caption{Micro-F on two settings (full and rolled-up ICDs) and for the four models when trained on MIMIC II or MIMIC III datasets.}
      \label{results}
    \end{table}

\begin{figure*}[t!]
   \centering
   \includegraphics[width=6.5in]{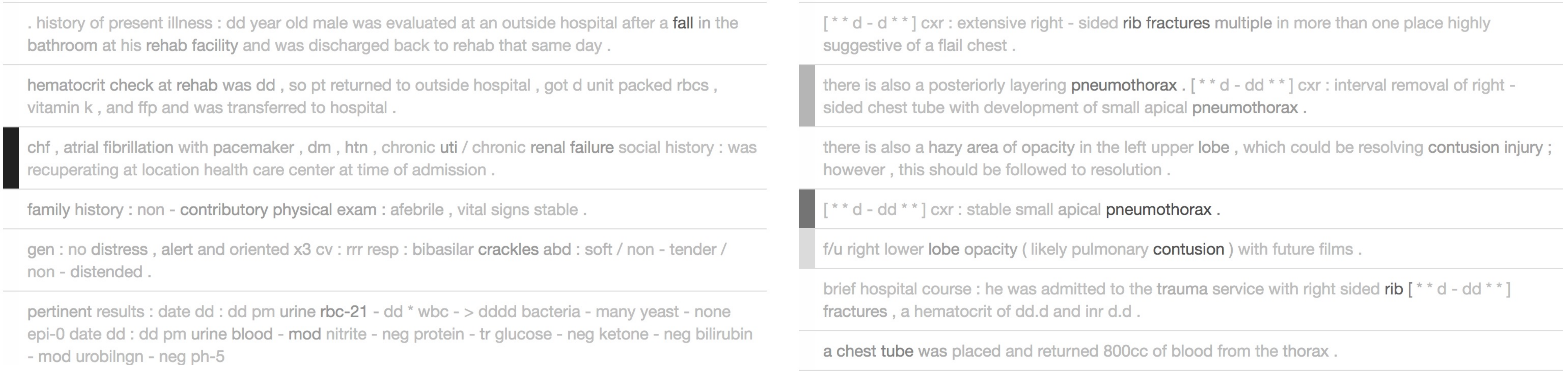}
   \caption{Sample text of a patient note (one sentence per line). On the left, visualization for the with attention weights at the sentence and word levels associated with the ICD9 codes, on the left sentence level attention weights for ICD9 code ``Heart failure", on the the right for code ``Traumatic pneumothorax and hemothorax".}\label{fig:visualizer}
\end{figure*}

\subsection{Label Frequency} 
\begin{figure}
   \centering
   \includegraphics[width=3in]{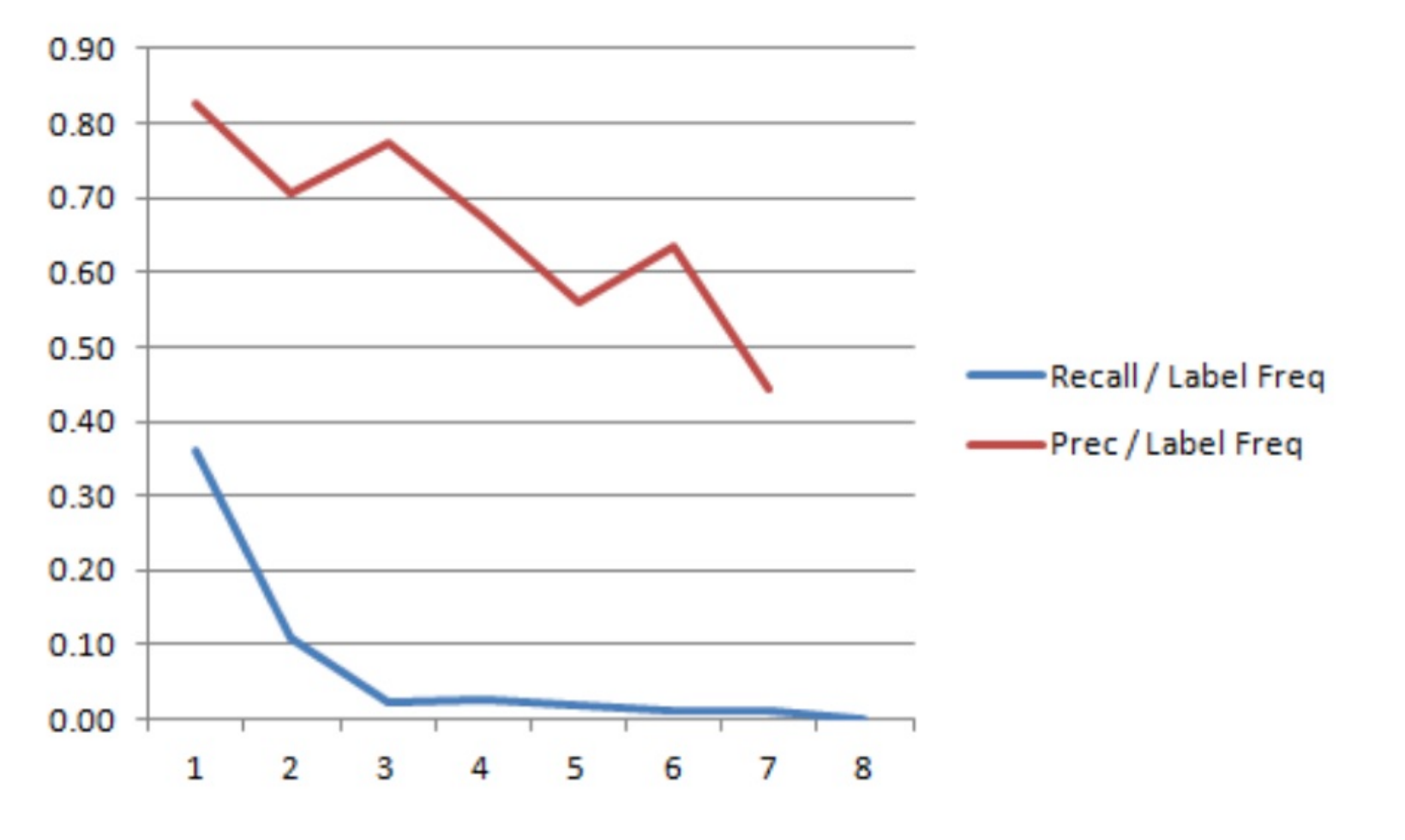}
   \caption{Effect label frequency on HA-GRU performance when trained on MIMIC III. X-axis represents the bins of labels ranked by their frequency in the training set.}\label{fig:freq}
\end{figure}

We also tested the effect label frequency on the performance of the HA-GRU classifier. We recalculated precision and recall scores on subsets of labels. The subsets were created by sorting the labels by frequency they appear in MIMIC-III dataset and binning them to groups of 50 labels. As such, bin 1 comprises the 50 most frequent ICD9 codes in the training set (with an average 12\% frequency over the records in the training set), codes in bin 2 had an average 1.9\% frequency, codes in bin 3 appeared in 1.1\% of the records, up to bin 8 which 0.2\% of the records in the training set. The effect can be seen in Figure \ref{fig:freq}. We note that the recall score drops much more dramatically than the precision as the label frequency decreases.

\subsection{Model Explaining Power}
We discuss how the CNN and HA-GRU architectures can support model explaining power. 
\paragraph*{CNN.} To analyze the CNN prediction we can test which n-grams triggered the max-pooling layer. Given a sentence with $n$ words we can feed it forward through the embedding layer and the convolution layer. The output of the convolution a list of vectors each the size of the number of channels of the convolution layer where vector corresponds to an n-gram. We can identify what triggered the max pooling layer by finding the maximum value of each channel.
Thus, for predicted labels, one of the activated n-grams does include information relevant for that label (whether correct for true positive labels or incorrect for false positive labels). For example in our experiments, for the label: \textit{``682.6-Cellulitis and abscess of leg, except foot''} one of the activated n-gram detected was \textit{``extremity cellulitis prior''}. 

This transparency process can also be useful for error analysis while building a model, as it can highlight True Positive and False Positive labels. However, it is difficult in the CNN to trace back the decisions for False Negatives predictions.

\paragraph*{HA-GRU}
For the HA-GRU model we can use attention weights to better understand what sentences and what words in that sentence contributed the most to each decision. We can find which sentence had the highest attention score for each label, and given the most important sentence, we can find what word received the highest attention score. For example, in our experiments for label \textit{``428-Heart failure''} we found that the sentence with the highest attention score was \textit{``d . congestive heart failure ( with an ejection fraction of dd \% to dd \% ) .''}, while the token \textit{``failure''} was found most relevant across all labels. Figure~\ref{fig:visualizer} provides additional examples. Note that the ``d" and ``dd" tokens are from the pre-procecssing step, which mapped all numbers to pseudo-tokens.

Like in the CNN, we can use this process for error analysis. In fact, the HA-GRU model explains prediction with greater precision, at the sentence level. For instance, we could explore the following False Positive prediction: the model assigned the label \textit{``331-Other cerebral degenerations''} to the sentence:\textit{``alzheimer 's dementia .''}. We can see that the condition was relevant to the medical note, but was mentioned under the patient's past medical history (and not a current problem). In fact, many of the False Positive labels under the HA-GRU model were due to mentions belonging to the past medical history section. This suggests that the coding task would benefit from a deeper architecture, with attention to discourse-level structure.

In contrast to the CNN, the HA-GRU model can also help analyze False Negative label assignments. When we explored the False Negative labels, we found that in many cases the model found a relevant sentence, but failed to classify correctly. This suggests the document-level attention mechanism is successful. For instance, for the False Negative \textit{``682-Other cellulitis and abscess''}, the most attended sentence was \textit{``... for right lower extremity cellulitis prior to admission ...''}. The false positive codes for this sentence included \textit{``250-Diabetes mellitus''} and \textit{``414-Other forms of chronic ischemic heart disease''}. We note that in the case of cellulitis, it is reasonable that the classifier preferred other, more frequent codes, as it is a common comorbid condition in the ICU.\footnote{Full visualizations of sample discharge summaries are provided at \url{https://www.cs.bgu.ac.il/~talbau/mimicdemo}}

\section{Conclusion}
We investigate four modern models for the task of extreme multi-label classification on the MIMIC datasets. Unlike previous work, we evaluate our models on all ICD9 codes thus making sure our models could be used for real world ICD9 tagging. The tokenization step, mapping rare variants using edit distance, improved results for CBOW and CNN models by \textasciitilde0.5\%, highlighting the importance of preprocessing data noise problems in real-world settings. 
The HA-GRU model not only achieves the best performance on the task of rolled-up codes (55.86\% $F1$ on MIMIC III, \textasciitilde2.8\% absolute improvement on the best SVM baseline) but is able to provide insight on the task for future work such as using discourse-level structure available in medical notes yet never used before. The ability to highlight the decision process of the model is important for adoption of such models by medical experts. 
On the sub-task of MIMIC II, which includes a smaller training dataset, HA-GRU achieved \textasciitilde7\% absolute $F1$ improvement, suggesting it requires less training data to achieve top performance, which is important for domain adaptation efforts when applying such models to patient records from other sources (such as different hospitals). 

\section{Acknowledgements}
This work is supported by National Institute of General Medical Sciences Grant R01GM114355 (NE) and Frankel Center for Computer Science .

\bibliography{aaai18}
\bibliographystyle{aaai}

\end{document}